\documentclass[runningheads]{llncs}

\usepackage{times}
\usepackage{soul}
\usepackage{url}
\usepackage[utf8]{inputenc}
\usepackage[small]{caption}
\usepackage{graphicx}
\usepackage{booktabs}
\usepackage{algorithm}
\usepackage{algorithmicx}
\usepackage{wrapfig,lipsum,booktabs}

\usepackage{float}
\usepackage{makecell}
\usepackage{times}
\usepackage{helvet}
\usepackage{courier}
\usepackage{graphicx}
\usepackage{amsfonts}
\usepackage{amsmath}
\usepackage{multirow}
\usepackage{todonotes}
\usepackage{xcolor}
\usepackage{algorithm}
\usepackage{algpseudocode}
\usepackage[title]{appendix}
\usepackage{hyperref}
\usepackage{apptools}
\AtAppendix{\renewtheorem{definition}{Definition}[section]
\renewtheorem{proposition}[definition]{Proposition}
\renewtheorem{lemma}[definition]{Lemma}}
    
\newcommand{\interp}{\mathcal{I}}     
\newcommand{\interpDom}{\Delta^\mathcal{I}}    
\newcommand{\interpMap}{\cdot^{\mathcal{I}}}  
\newcommand{\interpw}{\mathcal{I}_w}     
\newcommand{\interpDomw}{\Delta^{\interpw}}    
\newcommand{\interpMapw}{\cdot^{\interpw}}  
\newcommand{\abox}{\mathcal{A}}     
     
\newcommand{\tbox}{\mathcal{T}}     
            
\renewcommand{\Box}{\operatorname{Box}}
\newcommand{\Vol}{\operatorname{MVol}}

\newcommand{\contains}{\operatorname{Disjoint}}
\newcommand{\concept}{\textsf}
\newcommand{\rel}{\textsf}
\newcommand{\insta}{\textsf}

\title{ Faithful Embeddings for $\mathcal{EL}^{++}$ Knowledge Bases }
\titlerunning{BoxEL}

\author{%
Bo Xiong\inst{1} \and
Nico Potyka\inst{2}\and
Trung-Kien Tran\inst{3}\and
Mojtaba Nayyeri\inst{1}\and
Steffen Staab\inst{1,4}}

\institute{
University of Stuttgart, Stuttgart, Germany \\
\email{\{bo.xiong, mojtaba.nayyeri, steffen.staab\}@ipvs.uni-stuttgart.de} \and
Imperial College London, London, United Kingdom \\
\email{n.potyka@imperial.ac.uk}  \and
Bosch Center for Artificial Intelligence, Renningen, Germany \\
\email{trungkien.tran@de.bosch.com}  \and
University of Southampton, Southampton, United Kingdom\\
 }
 
\authorrunning{Bo Xiong et al.}

\begin{document}

\maketitle              

\begin{abstract}
Recently, increasing efforts are put into learning continual representations for symbolic knowledge bases (KBs). However, these approaches either only embed the data-level knowledge (ABox) or suffer from inherent limitations when dealing with concept-level knowledge (TBox), i.e., they cannot faithfully model the logical structure present in the KBs. 
We present BoxEL, a geometric KB embedding approach that allows for better capturing the logical structure (i.e., ABox and TBox axioms) in the description logic $\mathcal{EL}^{++}$.
BoxEL models concepts in a KB as axis-parallel \emph{boxes} that are suitable for modeling concept intersection, entities as points inside boxes, and relations between concepts/entities as \emph{affine transformations}. 
We show theoretical guarantees (\textit{soundness}) of BoxEL for preserving logical structure. Namely, the learned model of BoxEL embedding with loss $0$ is a (logical) model of the KB. Experimental results on (plausible) subsumption reasonings and a real-world application for protein-protein prediction show that BoxEL outperforms traditional knowledge graph embedding methods as well as state-of-the-art $\mathcal{EL}^{++}$ embedding approaches.

\end{abstract}

\section{Introduction}
Knowledge bases (KBs) provide a \textit{conceptualization} of objects and their relationships, which are of great importance in many applications like biomedical and intelligent systems~\cite{rector1996galen,gene2015gene}. 
KBs are often expressed using description logics (DLs)~\cite{baader2003description}, a family of languages allowing for expressing domain knowledge via logical statements (a.k.a axioms). 
These logical statements are divided into two parts: 1) an ABox consisting of \emph{assertions} over instances, i.e., factual statements like $\rel{isFatherOf}(\insta{John}, \insta{Peter})$;
2) a TBox consisting of logical statements constraining concepts, e.g., $\concept{Parent} \sqsubseteq \concept{Person}$.

KBs not only provide clear semantics in the application domains but also enable (classic) reasoners~\cite{DBLP:journals/ws/SteigmillerLG14,DBLP:journals/jar/KazakovKS14} to perform logical inference, i.e., making implicit knowledge explicit. Existing reasoners are highly optimized and scalable but they are limited to only computing classical logical entailment but not designed to perform inductive (analogical) reasoning and cannot handle noisy data. Embedding based methods, which map the objects in the KBs into a low dimensional vector space while keeping the similarity, have been proposed to complement the classical reasoners and shown remarkable empirical performances on performing (non-classical) analogical reasonings. 

Most KB embeddings methods~\cite{wang2017knowledge} focus on embedding data-level knowledge in ABoxes, a.k.a.,\ knowledge graph embeddings (KGEs). However, KGEs cannot preserve concept-level knowledge expressed in TBoxes. 
Recently, embedding methods for KBs expressed in DLs have been explored. Prominent examples include $\mathcal{EL}^{++}$ \cite{kulmanov2019embeddings} that supports conjunction and full existential quantification, and $\mathcal{ALC}$ \cite{ozccep2020cone} that further supports logical negation.
We focus on $\mathcal{EL}^{++}$, an underlying formalism of the OWL2 EL profile of the Web Ontology Language \cite{graua2008web}, which has been used in expressing various biomedical ontologies \cite{gene2015gene,rector1996galen}. For embedding $\mathcal{EL}^{++}$ KBs, several approaches such as Onto2Vec \cite{DBLP:journals/bioinformatics/SmailiGH18} and OPA2Vec \cite{smaili2019opa2vec} have been proposed. These approaches require annotation data and do not model logical structure explicitly. Geometric representations, in which the objects are associated with geometric objects such as balls \cite{kulmanov2019embeddings} and convex cones \cite{ozccep2020cone}, provide a high expressiveness on embedding logical properties. For $\mathcal{EL}^{++}$ KBs, ELEm \cite{kulmanov2019embeddings} represents concepts as open $n$-balls and relations as simple translations. Although effective, ELEm suffers from several major limitations: 

\begin{figure}[t!]
    \centering
    \includegraphics[width=0.9\linewidth]{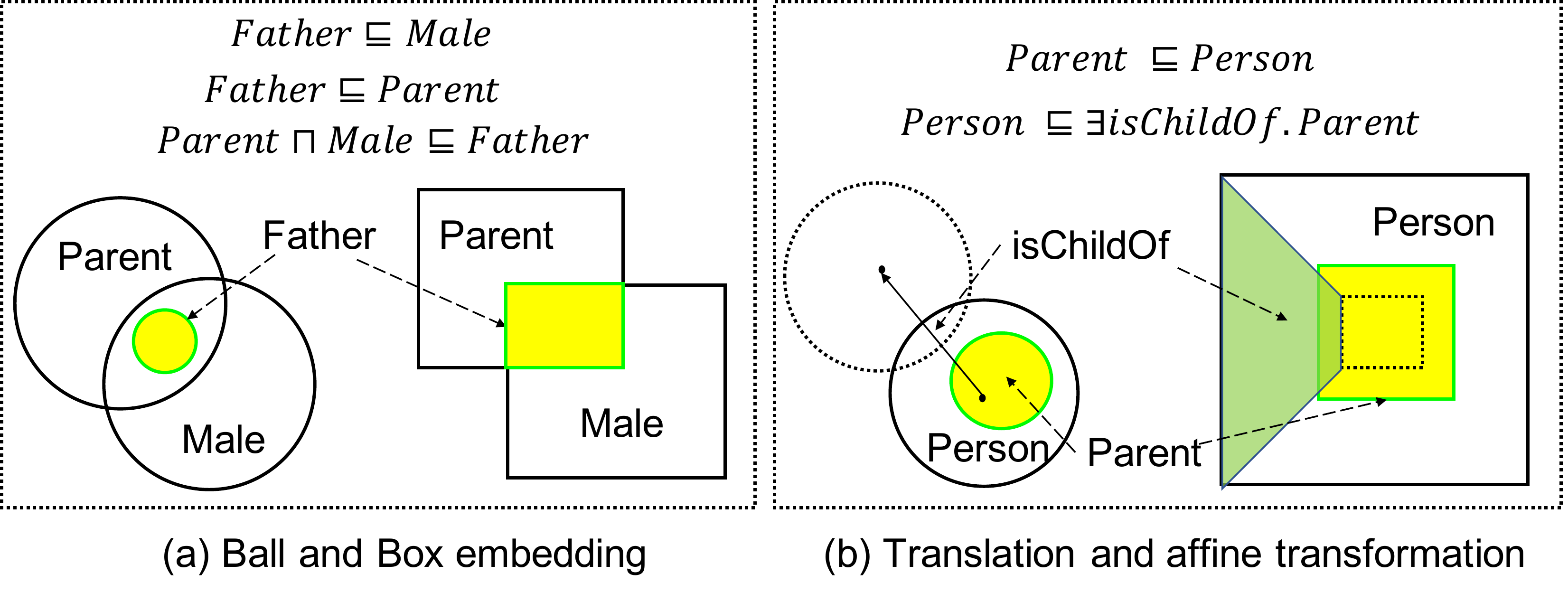}
    \caption{Two counterexamples of ball embedding and its relational transformation. (a) Ball embedding cannot express concept equivalence $\concept{Parent} \sqcap \concept{Male} \equiv \concept{Father}$ with intersection operator. (b) The \textit{translation} cannot model relation (e.g. $\rel{isChildOf}$) between $\concept{Person}$ and $\concept{Parent}$ when they should have different volumes. 
    These two issues can be solved by \textit{box} embedding and modelling relation as \textit{affine transformation} among boxes, respectively.}
    \label{fig:example_intersection}
\end{figure}

\begin{itemize}
    \item Balls are not closed under intersection and cannot faithfully represent concept intersections.
    For example, the intersection of two concepts $\concept{Parent} \sqcap \concept{Male}$, that is supposed to represent $\concept{Father}$, is not a ball (see Fig.~\ref{fig:example_intersection}(a)). 
    Therefore, the concept equivalence $\concept{Parent} \sqcap \concept{Male} \equiv \concept{Father}$ cannot be captured in the embedding space.
    \item The relational embedding with simple \textit{translation} causes issues for embedding concepts with varying sizes. For example, Fig.~\ref{fig:example_intersection}(b) illustrates the embeddings of the axiom  $\exists \rel{isChildOf}.\concept{Person} \sqsubseteq \concept{Parent}$ assuming the existence of another axiom $\concept{Parent} \sqsubseteq \concept{Person}$. In this case, it is impossible to \textit{translate} the larger concept $\concept{Person}$ into the smaller one $\concept{Parent}$,\footnote{Under the \textit{translation} setting, the embeddings will simply become $\concept{Parent} \equiv \concept{Person}$, which is obviously not what we want as we can express $\concept{Parent} \not\equiv \concept{Person}$ with $\mathcal{EL}^{++}$ by propositions like $\concept{Children} \sqcap \concept{Parent} \sqsubseteq \bot$, $\concept{Children} \sqsubseteq \concept{Person}$ and $\concept{Children}(a)$} as it does not allow for scaling the size.
    \item ELEm does not distinguish between entities in ABox and concepts in TBox, but rather regards ABox axioms as special cases of TBox axioms. This simplification cannot fully express the logical structure, e.g., an entity must have minimal volume. 
\end{itemize}

To overcome these limitations, we consider modeling concepts in the KB as \textit{boxes} (i.e., axis-aligned hyperrectangles), encoding entities as points inside the boxes that they should belong to, and the relations as the \textit{affine transformation} between boxes and/or points. Fig.~\ref{fig:example_intersection}(a) shows that the box embedding has closed form of intersection and the \textit{affine transformation} (Fig.~\ref{fig:example_intersection}(b)) can naturally capture the cases that are not possible in ELEm. 
In this way, we present BoxEL for embedding $\mathcal{EL}^{++}$ KBs, in which the interpretation functions of $\mathcal{EL}^{++}$ theories in the KB can be represented by the geometric transformations between boxes/points. 
We formulate BoxEL as an optimization task by designing and minimizing various loss terms defined for each logical statement in the KB. 
We show theoretical guarantee (\textit{soundness}) of BoxEL in the sense that if the loss of BoxEL embedding is $0$, then the trained model is a (logical) model of the KB.
Experiments on (plausible) subsumption reasoning over three ontologies and predicting protein-protein interactions show that BoxEL outperforms previous approaches. 

\section{Related Work}

Knowledge graph embeddings (KGEs) have been developed for different tasks. Early works, which focus on link prediction, embed both entities and relations as vectors in a vector space to model the relationships between entities~\cite{BordesUGWY13,TrouillonWRGB16,DettmersMS018}. Prominent examples include \emph{additive} (or \emph{translational}) family \cite{BordesUGWY13,wang2014knowledge,DBLP:conf/aaai/LinLSLZ15} and \emph{multiplicative} (or \emph{bilinear}) family \cite{DBLP:conf/icml/NickelTK11,yang2014embedding,DBLP:conf/icml/LiuWY17}. 
Such techniques only embed the data-level part of KBs and work relatively well for the link prediction tasks. However, KGEs demonstrate limitations when being used to learn the representation of background knowledge such as ontologies of logical rules~\cite{garg2019quantum,ozccep2020cone}, as well as complex logical query~\cite{ren2020query2box,betae}.

Inspired by the theory of conceptual spaces \cite{DBLP:books/daglib/0006106}, several methods have been proposed to embed concepts as convex regions in vector spaces \cite{DBLP:conf/kr/Gutierrez-Basulto18}, including balls ~\cite{kulmanov2019embeddings} and convex cones \cite{ozccep2020cone}.  
Such conceptual/geometric methods nicely model the set-theoretic semantics that can be used to capture logical rules of knowledge graphs~\cite{BoxE}, transitive closure in graphs~\cite{vilnis2018probabilistic} and logical query for multi-hop question answering~\cite{ren2020query2box}. 

Among embeddings for complex concept descriptions, 
boxes have some conceptual advantages, but they have 
not been exploited for representing ontologies yet. 
BoxE \cite{BoxE} does embed some logical rules but mostly focus on embedding the relational patterns in ABoxes. 
In contrast, our approach BoxEL focuses on $\mathcal{EL}^{++}$ that has a larger TBox and  provides soundness guarantees. BoxEL is closely related to ELEm~\cite{kulmanov2019embeddings}, but instead of using ball embedding and translation, we consider box embedding and affine transformation that have various inherent advantages as discussed before. Another difference of our method is that we use a different encoding that distinguishes between entities and concepts. 
Furthermore, we take advantage of the volume of boxes for disjointedness representation, resulting in a more natural encoding of the disjointedness of concepts, i.e., two concepts are disjoint iff their intersection has zero volume. 

\begin{wraptable}{r}{0.58\linewidth}
\vspace{-0.4cm}
\centering
\caption{Syntax and semantic of $\mathcal{EL}^{++}$ (role inclusions and concrete domains are omitted).}
\label{tab:el_syntax_seman}
\resizebox{\linewidth}{!}{
\begin{tabular}{ccccccc}
\hline\noalign{\smallskip} 
 & Name & Syntax & Semantics \\
\hline\noalign{\smallskip}
\multirow{5}{*}{Constructors} & Top concept &  $\top$ & $\Delta^{\mathcal{I}}$ \\
& Bottom concept & $\bot$ & $\emptyset$ \\
& Nominal & $\{a\}$ & $\{a^{\mathcal{I}}\}$ \\
& Conjunction & $C\sqcap D$ & $C^{\mathcal{I}}\cap D^{\mathcal{I}}$ \\
& Existential restriction & $\exists r . C$ & \makecell{$\left\{x \in \Delta^{\mathcal{I}} \mid \exists y \in \Delta^{\mathcal{I}}\right.$ \\
$\left.(x, y) \in r^{\mathcal{I}} \wedge y \in C^{\mathcal{I}}\right\}$ } \\
\hline\noalign{\smallskip}
\multirow{2}{*}{ABox}
& Concept assertion & $C(a)$ & $a^{\mathcal{I}} \in C^{\mathcal{I}} $ \\
& Role assertion & $r(a,b)$ & $(a^{\mathcal{I}}, b^{\mathcal{I}}) \in r^{\mathcal{I}}$ \\
\hline\noalign{\smallskip}
\multirow{1}{*}{TBox}
& Concept inclusion & $C \sqsubseteq D$ & $C^{\mathcal{I}} \subseteq D^{\mathcal{I}}$  \\
\noalign{\smallskip}\hline
\end{tabular}
}
\end{wraptable}



\section{Description Logic $\mathcal{EL}^{++}$}

We consider the DL $\mathcal{EL}^{++}$ that underlies multiple biomedical KBs like GALEN \cite{rector1996galen} and  the Gene Ontology \cite{gene2015gene}. Formally, the syntax of $\mathcal{EL}^{++}$ is built up from a set $N_I$ of \emph{individual names}, $N_C$ of \emph{concept names} and $N_R$ of \emph{role names} (also called \emph{relations}) using the constructors shown in Table \ref{tab:el_syntax_seman}, where $N_I$, $N_C$ and $N_R$ are pairwise disjoint. 

The semantics of $\mathcal{EL}^{++}$ is defined by \emph{interpretations} $\interp = (\interpDom, \interpMap)$, where the domain $\interpDom$ is a non-empty set
and $\interpMap$ is a mapping that associates every individual with an element in $\interpDom$, every concept name with a subset of $\interpDom$, and every relation name with a relation over $\interpDom \times \interpDom$. An \emph{interpretation} is satisfied if it satisfies the corresponding semantic conditions. The syntax and the corresponding semantics (i.e., interpretation of concept expressions) of $\mathcal{EL}^{++}$ are summarized in Table \ref{tab:el_syntax_seman}. 



An $\mathcal{EL}^{++}$ KB $(\abox, \tbox)$ consists of an ABox $\abox$ and a TBox $\tbox$. The \emph{ABox} is a set of \emph{concept assertions} ($C(a)$)  and \emph{role assertions}  ($r(a,b)$), where $C$ is a concept, $r$ is a relation, and $a,b$ are individuals. 
The TBox is a set of \emph{concept inclusions} of the form $C \sqsubseteq D$.
Intuitively, the ABox contains instance-level information (e.g. $\concept{Person}(\insta{John})$), $\rel{isFatherOf}(\insta{John}, \insta{Peter})$), while the TBox contains information about concepts (e.g. $\concept{Parent} \sqsubseteq \concept{Person}$ ). 
Every $\mathcal{EL}^{++}$ KB can be transformed
such that every TBox statement has the form $C_1 \sqsubseteq D$, $C_1 \sqcap C_2 \sqsubseteq D$, $C_1 \sqsubseteq \exists r. C_2$, 
$\exists r. C_1 \sqsubseteq D$, where $C_1, C_2, D$ can be the top 
concept, concept names or nominals and $D$ can also be the bottom concept \cite{baader2005pushing}. 
The normalized KB can be computed in linear time by introducing new concept names for complex concept expressions and is a conservative extension of the original KB, i.e., every model of the normalized KB is a model of the original KB and every model of the original KB can be extended to be a model
of the normalized KB \cite{baader2005pushing}.

\section{BoxEL for Embedding $\mathcal{EL}^{++}$ Knowledge Bases}
In this section, we first present the geometric construction process of $\mathcal{EL}^{++}$ with box embedding and affine transformation, followed by a discussion of the geometric interpretation. Afterward, we describe the BoxEL embedding by introducing proper loss function for each ABox and TBox axiom. Finally, an optimization method is described for the training of BoxEL. 

\subsection{Geometric Construction} 
We consider a KB $(\abox, \tbox)$ consisting
of an ABox $\abox$ and a TBox $\tbox$ where $\tbox$ has been normalized as explained before. 
Our goal is to associate entities (or individuals) with points and concepts with boxes in $\mathbb{R}^n$ such that the axioms in the KB are respected.

To this end, we consider two functions $m_w, M_w$ parameterized
by a parameter vector $w$ that has to be learned.
Conceptually, we consider points as boxes of volume $0$. This will be helpful later to encode the meaning of axioms for points and boxes in a uniform way.
Intuitively, $m_w: N_I \cup N_C \rightarrow \mathbb{R}^n$ maps individual and concept names to the lower left corner and  $M_w: N_I \cup N_C \rightarrow \mathbb{R}^n$ maps
them to the upper right corner of the box that represents them.
For individuals $a \in N_I$, we have $m_w(a) = M_w(a)$, so that it is sufficient 
to store only one of them.
The \emph{box associated with $C$} is defined as 
\begin{equation}
    \Box_w(C) = \{x \in \mathbb{R}^n \mid m_w(C) \leq x \leq M_w(C) \},
\end{equation}
where the inequality is defined component-wise. 

Note that boxes are closed under intersection, which allows us to compute the volume of the intersection of boxes. The lower corner of the box $\Box_w(C) \cap \Box_w(D)$
is $\max(m_w(C), m_w(D))$ and the upper corner is 
$\min\left(M_w(C), M_w(D)\right)$, where minimum and maximum are taken component-wise.
The volume of boxes can be used to encode axioms in a very concise way. 
However, as we will describe later, one problem is that points have volume $0$. This does not allow distinguishing empty boxes from points. To show that our encoding
correctly captures the logical meaning of axioms, we will consider a 
\emph{modified volume} that assigns a non-zero volume to points and some empty boxes. 
The (modified) volume of a box is defined as
\begin{equation}\label{eq:modi_vol}
    \Vol(\Box_w(C)) = \prod_{i=1}^n \max(0, M_w(C)_i - m_w(C)_i + \epsilon),
\end{equation}
where $\epsilon>0$ is a small constant. A point now
has volume $\epsilon^n$. Some empty boxes can actually have arbitrarily large modified volume.
For example the 2D-box with lower corner $(0,0)$ and upper corner $(-\frac{\epsilon}{2},N)$ has volume $\frac{\epsilon\cdot N}{2}$. 
While this is not meaningful geometrically, it does not cause any problems for our encoding because we only want to ensure that boxes with zero volume are empty (and not points). In practice, we will use \textit{softplus volume} as approximation (see Sec. \ref{sec:soft}). 

We associate every role name $r \in N_r$ with an affine transformation denoted by
$T^r_w(x) = D^r_w x + b^r_w$, where $D^r_w$ is an $(n \times n)$ diagonal matrix with non-negative entries and $b^r_w \in \mathbb{R}^n$ is a vector. 
In a special case where all diagonal entries of $D^r_w$ are $1$, $T^r_w(x)$ captures translations. 
Note that relations have been represented
by translation vectors analogous to TransE in \cite{kulmanov2019embeddings}. However, this necessarily means
that the concept associated with the range of a role has the same size
as its domain. This does not seem very intuitive, in particular, for
N-to-one relationships like $has\_nationality$ or $lives\_in$ that map
many objects to the same object. 
Note that $T^r_w( \Box_w(C)) = \{T^r_w(x) \mid x \in \Box_w(C)\}$ is the box
with lower corner $T^r_w( m_w(C))$ and upper corner  $T^r_w( M_w(C))$.
To show this, note that $m_w(C) < M_w(C)$ implies
$D^r_w m_w(C)\leq D^r_w M_w(C)$ because $D^r_w$ is a diagonal matrix with
non-negative entries. Hence,
$T^r_w( m_w(C)) = 
D^r_w m_w(C) + b^r_w
\leq D^r_w M_w(C) + b^r_w =
T^r_w( M_w(C))$. For $m_w(C) \geq M_w(C)$, both 
$\Box_w(C)$ and 
$T^r_w( \Box_w(C))$ are empty.

Overall, we have the following parameters:
\begin{itemize}
    \item for every individual name $a \in N_I$, we have $n$ parameters  
    for the vector $m_w(a)$ (since $m_w(a)=M_w(A)$, we have to store only one
    of $m_w$ and $M_w$),
    \item for every concept name $C \in N_C$, we have $2n$ parameters for
    the vectors $m_w(C)$ and $M_w(C)$,
    \item for every role name $r \in N_r$, we have $2n$ parameters. $n$ parameters for the diagonal elements of $D^r_w$ and $n$ parameters
    for the components of $b^r_w$.
\end{itemize}
As we explained informally before, $w$ summarizes all parameters.
The overall number of parameters in $w$ is
$n \cdot (|N_I| + 2 \cdot |N_C| +  2 \cdot |N_r|)$.

\begin{figure}[t!]
    \centering
    \includegraphics[width=0.9\linewidth]{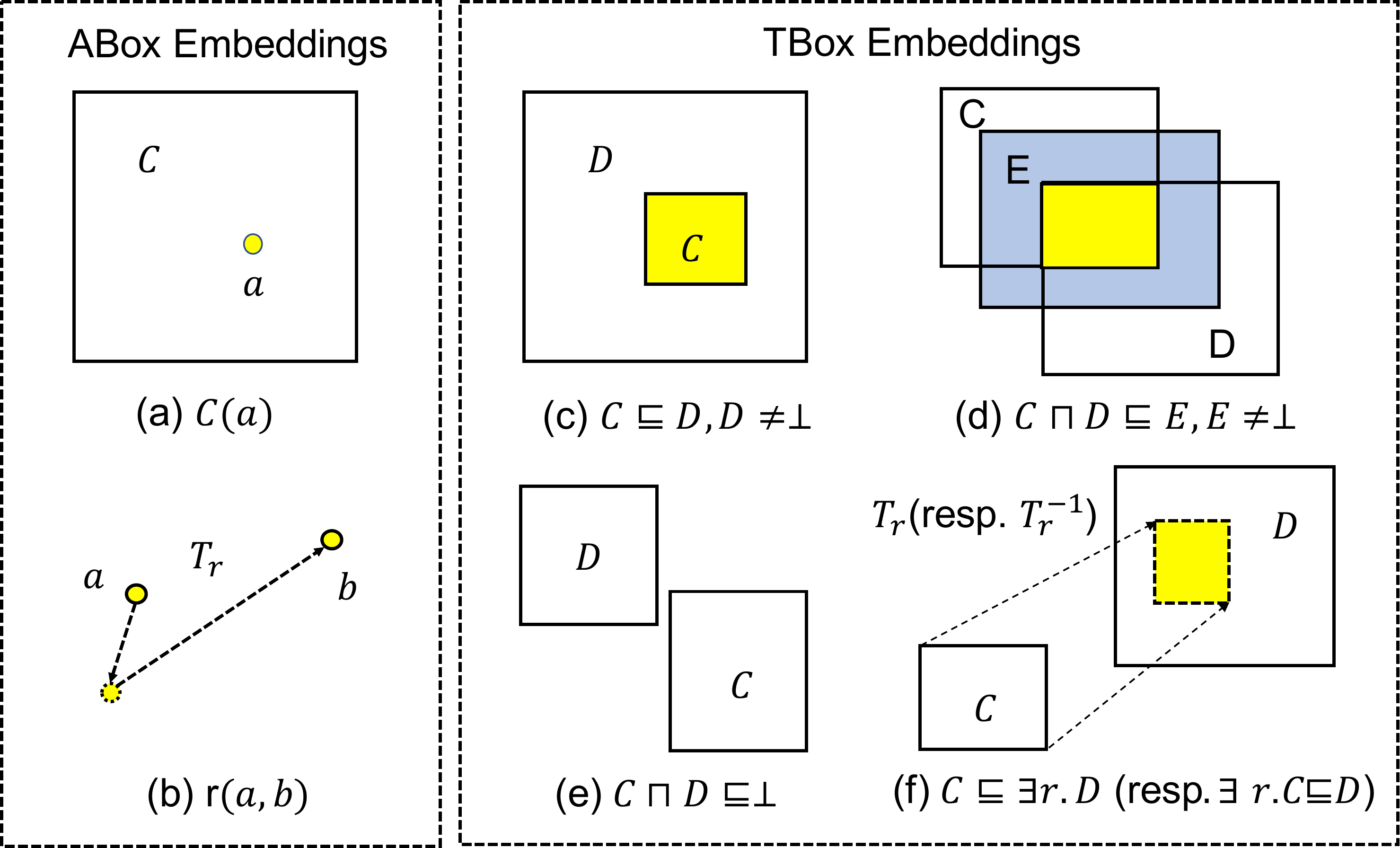}
    \caption{The geometric interpretation of logical statements in ABox (\textit{left}) and TBox (\textit{right}) expressed by DL $\mathcal{EL}^{++}$ with BoxEL embeddings. The concepts are represented by \textit{boxes}, entities are represented by \textit{points} and relations are represented by \textit{affine transformations}. $T_r$ and $T_r^{-1}$ denote the transformation function of relation $r$ and its inverse function, respectively.}
    \label{fig:box_el}
\end{figure}

\subsection{Geometric Interpretation}

The next step is to encode the axioms in our KB.
However, we do not want to do this in an arbitrary fashion, but, ideally, 
in a way that gives us some analytical guarantees.
\cite{kulmanov2019embeddings} made an interesting first step by showing that their encoding is \emph{sound}.
In order to understand soundness, it is important to know that the parameters 
of the embedding are learnt by minimizing a loss function that contains a 
loss term for every axiom. Soundness then means that if the loss function
yields $0$, then the KB is satisfiable. Recall that satisfiability means
that there is an interpretation that satisfies all axioms in the KB. Ideally, we should be able to construct such an interpretation directly
from our embedding. This is indeed what the authors in \cite{kulmanov2019embeddings} did. The idea is that points in the vector
space make up the domain of the interpretation, the points that lie in regions
associated with concepts correspond to the interpretation of this concept
and the interpretation of roles correspond to translations between points
like in TransE. In our context, geometric interpretation
can be defined as follows.

\begin{definition}[Geometric Interpretation]
Given a parameter vector $w$ representing 
an $\mathcal{EL}^{++}$ embedding, the corresponding \emph{geometric interpretation}  $\interpw = (\interpDomw, \interpMapw)$ is defined 
as follows:
\begin{enumerate}
    \item $\interpDomw = \mathbb{R}^n$,
    \item for every concept name $C \in N_C$, $C^{\interpw} =\Box_w(C)$,
    \item for every role $r \in N_R$, $r^{\interpw} = \{(x, y) \in \interpDomw \times \interpDomw \mid T^r_w(x) = y \}$,
    \item for every individual name $a \in N_I$, $a^{\interpw} = m_w(a)$.
\end{enumerate}
\end{definition}
We will now encode the axioms by designing one loss term for every axiom in a normalized $\mathcal{EL}^{++}$ KB, such that the axiom is satisfied by the geometric interpretation when the loss is $0$. 

\subsection{ABox Embedding}
ABox contains concept assertions and role assertions. We introduce the following two loss terms that respect the geometric interpretations. 
\subsubsection{Concept Assertion} Geometrically, a concept assertion $C(a)$ asserts that the point $m_w(a)$ is inside the box $\Box_w(C)$ (see Fig.~\ref{fig:box_el}(a)). This can be expressed by demanding
$m_w(C) \leq m_w(a) \leq M_w(C)$ for every component. 
The loss $\mathcal{L}_{C(a)}(w)$ is defined by
\begin{align}
\mathcal{L}_{C(a)}(w) = & \sum_{i=1}^n \left\|\max(0, m_w(a)_i- M_w(C)_i)\right\|_2 \notag + \sum_{i=1}^n \left\|\max(0, m_w(C)_i- m_w(a)_i)\right\|_2.
\end{align}

\subsubsection{Role Assertion} Geometrically, a role assertion $r(a,b)$ means that the point $m_w(a)$ should be mapped to $m_w(b)$ by the transformation $T^r_w$ (see Fig.~\ref{fig:box_el}(b)). That is, we should have $T^r_w(m_w(a))=m_w(b)$. 
We define a loss term
\begin{equation}
    \mathcal{L}_{r(a,b) }(w) =\left\|T^r_w(m_w(a)) - m_w(b) \right\|_2.
\end{equation}
It is clear from the definition that when the loss terms are $0$, the axioms are satisfied in their geometric interpretation.
\begin{proposition}\label{prop:1}
We have 
\begin{enumerate}
    \item If $\mathcal{L}_{C(a)}(w) = 0$, then $\interpw \models C(a)$,
    \item If $\mathcal{L}_{r(a,b)}(w)=0$, then $\interpw \models r(a,b)$.
\end{enumerate}
\end{proposition}

\subsection{TBox Embedding}
For the TBox, we define loss terms for the four cases in
the normalized KB. Before doing so, we define an auxiliary function
that will be 
used inside these loss terms.
\begin{definition}[Disjoint measurement]
Given two boxes $B_1, B_2$, the disjoint measurement can be defined by the (modified) volumes of $B_1$ and the intersection box $B_1 \cap B_2$,
\begin{equation}
   \contains(B_1, B_2) = 1-\frac{\Vol(B_1 \cap B_2)}{\Vol(B_1)}. 
\end{equation}
\end{definition}
We have the following guarantees.
\begin{lemma}
\label{contains_lemma}
\begin{enumerate}
    \item $0 \leq \contains(B_1, B_2) \leq 1$,
    \item $\contains(B_1, B_2) = 0$ implies $B_1 \subseteq B_2$,
    \item $\contains(B_1, B_2) = 1$ implies $B_1 \cap B_2 = \emptyset$.
\end{enumerate}
\end{lemma}


\subsubsection{NF1: Atomic Subsumption} An axiom of the form $C \sqsubseteq D$ geometrically means that $\Box_w(C) \subseteq \Box_w(D)$ (see Fig.~\ref{fig:box_el}(c)). 
If $D\neq \bot$, we consider the loss term
\begin{equation}\label{eq:loss_nf1}
    \mathcal{L}_{C \sqsubseteq D}(w) = \contains(\Box_w(C), \Box_w(D)).
\end{equation}
For the case $D=\bot$ where $C$ is not a nominal, e.g., $C \sqsubseteq \bot$, we define the loss term 
\begin{equation}
    \mathcal{L}_{C \sqsubseteq \bot}(w) = 
    \max(0, M_w(C)_0 - m_w(C)_0 + \epsilon).
\end{equation}
If $C$ is a nominal, the axiom is inconsistent and our model can just return an error.
\begin{proposition}
If $\mathcal{L}_{C \sqsubseteq D}(w)=0$, 
then $\interpw \models C \sqsubseteq D$,
where we exclude the inconsistent case $C=\{a\}, D=\bot$.
\end{proposition}

\subsubsection{NF2: Conjunct Subsumption} An axiom of the form 
$C \sqcap D \sqsubseteq E$ means that $\Box(C) \cap \Box(D) \subseteq \Box(E)$
(see Fig.~\ref{fig:box_el}(d)).
Since $\Box(C) \cap \Box(D)$ is a box again, we can use the same idea as
for NF1. For the case $E\neq \bot$, we define the loss term as
\begin{equation}\label{eq:loss_nf2}
    \small
    \mathcal{L}_{C \sqcap D \sqsubseteq E}(w) = 
    \contains(\Box_w(C) \cap \Box_w(D), \Box_w(E)).
\end{equation}
For $E = \bot$, the axiom states that $C$ and $D$ must be disjoint. The disjointedness can be interpreted as the volume of the intersection of the associated boxes being $0$ (see Fig.~\ref{fig:box_el}(e)). 
However, just using the volume as a loss term may not work well
because a minimization algorithm may minimize the volume of the boxes
instead of the volume of their intersections. Therefore, we normalize the
loss term by dividing by the volume of the boxes. Given by
\begin{equation}\label{eq:loss_disjoint}
    \small
    \mathcal{L}_{C \sqcap D \sqsubseteq \bot}(w) = \frac{\Vol(\Box_w(C)\cap \Box_w(D))}{\Vol(\Box_w(C)) + \Vol(\Box_w(D))}.
\end{equation}
\begin{proposition}
If $\mathcal{L}_{C \sqcap D \sqsubseteq E}(w)=0$,
then $\interpw \models C \sqcap D \sqsubseteq E$,
where we exclude the inconsistent case ${a} \sqcap {a} \sqsubseteq \bot$ (that is, $C=D=\{a\}, E=\bot$).
\end{proposition}

\subsubsection{NF3: Right Existential}
Next, we consider axioms of the form $C \sqsubseteq \exists r.D$.
Note that $\exists r.D$ describes those entities that are in  
relation $r$ with an entity from $D$.
Geometrically, those are points that are mapped to points in
$\Box_w(D)$ by the affine transformation corresponding to $r$. 
$C \sqsubseteq \exists r.D$ then means that every point in
$\Box_w(C)$ must be mapped to a point in $\Box_w(D)$, that is
the mapping of $\Box_w(C)$ is contained in $\Box_w(D)$ (see Fig.~\ref{fig:box_el}(f)).
Therefore, the encoding comes again down to encoding a
subset relationship as before. The only difference to the first 
normal form is that $\Box_w(C)$ must be mapped by the affine
transformation $T^r_w$. These considerations lead
to the following loss term
\begin{equation}\label{eq:loss_nf3}
    \mathcal{L}_{C \sqsubseteq \exists r.D}(w) =
    \contains(T^r_w(\Box_w(C)), \Box_w(D)).
\end{equation}
\begin{proposition}
If $\mathcal{L}_{C \sqsubseteq \exists r.D}(w) = 0$, 
then $\interpw \models C \sqsubseteq \exists r.D$.
\end{proposition}
\par

\subsubsection{NF4: Left Existential}
Axioms of the form $\exists r.C \sqsubseteq D$ can be treated
symmetrically to the previous case (see Fig.~\ref{fig:box_el}(f)). We only consider the case $D\neq \bot$ and define the loss
\begin{equation}\label{eq:loss_nf4}
    \mathcal{L}_{\exists r.C \sqsubseteq D }(w) =
    \contains(T^{-r}_w(\Box_w(C)), \Box_w(D)),
\end{equation}
where $T^{-r}_w$ is the inverse function of $T^{r}_w$ that is defined by
$T^{-r}_w(x) =  D^{-r}_w x - D^{-r}_w b^r_w$, where $D^{-r}_w$ is 
obtained from $D^{r}_w$ by replacing all diagonal elements with  their
reciprocal. Strictly speaking, the inverse only exists 
if all diagonal entries of $D^{r}_w$ are non-zero. 
However, we assume that the entries that occur in a loss
term of the form $\mathcal{L}_{\exists r.C \sqsubseteq D }(w)$
remain non-zero in practice when we learn them iteratively. 
\begin{proposition}
If $\mathcal{L}_{\exists r.C \sqsubseteq D }(w) = 0$, 
then $\interpw \models \exists r.C \sqsubseteq D$.
\end{proposition}

\subsection{Optimization}

\subsubsection{Softplus Approximation}\label{sec:soft}
For optimization, while the computation of the volume of boxes is straightforward,
using a precise \emph{hard volume} is known to cause problems when learning the parameters using gradient descent algorithms, e.g. there is no training signal (gradient flow) when box embeddings that should overlap but become disjoint \cite{li2018smoothing,patel2020representing,dasgupta2020improving}.
To mitigate the problem, we approximate the volume of boxes by the \textit{softplus volume} \cite{patel2020representing} due to its simplicity.
\begin{equation}\small\label{eq:softplus_volume}
 \operatorname{SVol}\left( \Box_w\left(C\right)\right)= \prod_{i=1}^n \operatorname{Softplus}_{t} \left( M_w\left(C  \right)_i - m_w\left(C\right)_i\right)
\end{equation}
where $t$ is a temperature parameter. The softplus function is defined as $\operatorname{softplus}_{t}(x)=t \log \left(1+e^{x/t}\right)$, which can be regarded as a smoothed version of the ReLu function ($\max\{0,x\}$) used for calculating the volume of \textit{hard boxes}. 
In practice, the \textit{softplus volume} is used to replace the \textit{modified volume} in Eq.(\ref{eq:modi_vol}) as it empirically resolves the same issue that point has zero volume. 

\subsubsection{Regularization} We add a regularization term in Eq.(\ref{eq:regularizer}) to all non-empty boxes to encourage that the boxes lie in the unit box $[0,1]^n$. 
\begin{equation}\label{eq:regularizer}
    \lambda = \sum_{i=1}^n \max(0, M_w(C)_i -1 + \epsilon) + \max(0, -m_w(C)_i-\epsilon)
\end{equation}
In practice, this also avoids numerical stability issues. For example, to minimize a loss term, a box that should have a fixed volume could become very \textit{slim}, i.e. some side lengths be extremely large while others become extremely small.

\subsubsection{Negative Sampling} In principle, the embeddings can be optimized without negatives. However, we empirically find that the embeddings will be highly overlapped without negative sampling. e.g. for role assertion $r(a,b)$, $a$ and $b$ will simply become the same point. We generate negative samples for the role assertion $r(a,b)$ by randomly replacing one of the head or tail entity. 
Finally, we sum up all the loss terms, and learn the embeddings by minimizing the loss with Adam optimizer \cite{DBLP:journals/corr/KingmaB14}. 

\section{Empirical Evaluation}

\subsection{A Proof-of-concept Example}
We begin by first validating the model in modeling a toy ontology--family domain \cite{kulmanov2019embeddings}, which is described by the following axioms:\footnote{Compared with the example given in \cite{kulmanov2019embeddings}, we add additional concept assertion statements that distinguish entities and concepts:}
\begin{equation*}\label{eq:family_domain}
\footnotesize
\begin{array}{cl}
\concept {Male} \sqsubseteq \concept {Person} & \concept{Female} \sqsubseteq \concept { Person } \\
\concept { Father } \sqsubseteq \concept { Male } & \concept { Mother } \sqsubseteq \concept { Female } \\
\concept { Father } \sqsubseteq \concept { Parent } & \concept {Mother} \sqsubseteq \concept { Parent }\\
\concept { Female } \sqcap \concept { Male } \sqsubseteq \bot & \concept{ Female } \sqcap \concept { Parent } \sqsubseteq \concept { Mother } \\
\concept { Male } \sqcap \concept { Parent } \sqsubseteq \concept { Father } & \exists \rel {hasChild.Person } \sqsubseteq \concept { Parent }\\
\concept { Parent } \sqsubseteq \concept { Person } & \concept { Parent } \sqsubseteq \exists \rel{ hasChild.Person } \\
\concept{Father}(\text{Alex}) & \concept{Father}(\text{Bob}) \\
\concept{Mother}(\text{Marie}) & \concept{Mother}(\text{Alice})
\end{array}
\end{equation*}

\begin{wrapfigure}{r}{0.53\linewidth}
\centering
    \includegraphics[width=\linewidth]{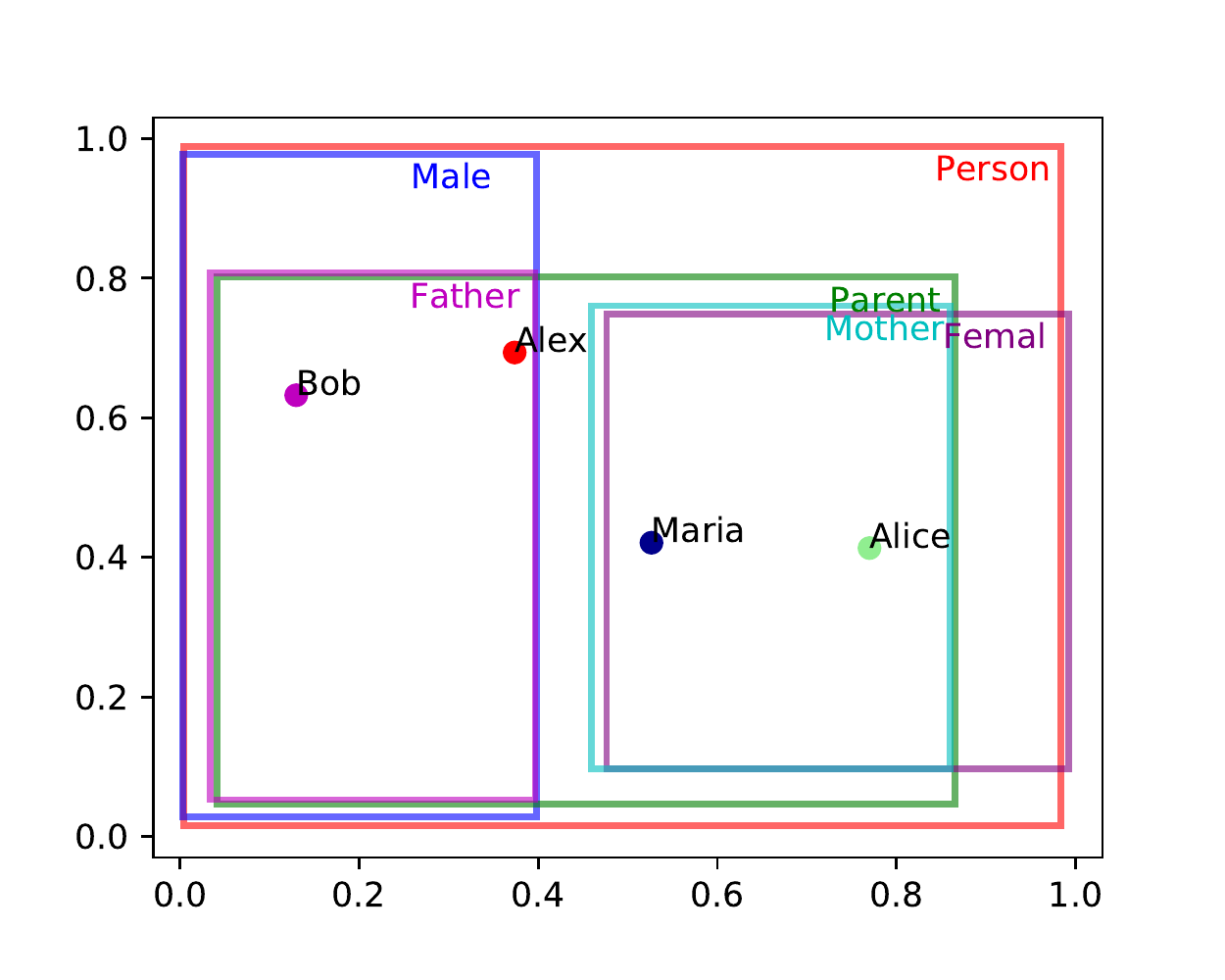}
    \caption{BoxEL embeddings in the family domain.
    }\label{fig:toy_example}
\end{wrapfigure}

We set the dimension to $2$ to visualize the embeddings. Fig.~\ref{fig:toy_example} shows that the generated embeddings accurately encode all of the axioms.
In particular, the embeddings of $\concept{Father}$ and $\concept{Mother}$ align well with the conjunction $\concept{Parent} \sqcap \concept{Male}$ and $\concept{Parent} \sqcap \concept{Female}$, respectively, which is impossible to be achieved by ELEm.


\subsection{Subsumption Reasoning}\label{app:subsumption}
We evaluate the effectiveness of BoxEL on (plausible) subsumption reasoning (also known as ontology completion). The problem is to predict whether a concept is subsumed by another one.
For each subsumption pair $C \sqsubseteq D$, the scoring function can be defined by 
\begin{equation}
    P(C \sqsubseteq D) = \frac{\Vol(\Box(C) \cap \Box(D))}{\Vol(\Box(C))}.
\end{equation}
Note that such subsumption relations are not necessary to be 
(logically)
entailed by the input KB, e.g., a subsumption relation can be plausibly inferred by $P(C \sqsubseteq D)=0.9$, allowing for non-classical plausible reasoning.
While the subsumption reasoning does not need negatives, we add an additional regularization term for non-subsumption axiom. In particular, for each atomic subsumption axiom $C \sqsubseteq D$, we generate a non-subsumption axiom $C \not\sqsubseteq D^\prime$ or $C^\prime \not\sqsubseteq D$ by randomly replacing one of the concepts $C$ and $D$. Note that this does not produce regular negative samples as the generated concepts pair does not have to be disjoint. Thus, the loss term for non-subsumption axiom cannot be simply defined by $\mathcal{L}_{C \not\sqsubseteq D^\prime} = 1-\mathcal{L}_{C \sqsubseteq D^\prime}$. Instead, we define the loss term as $\mathcal{L}_{C \not\sqsubseteq D^\prime} = \phi(1-\mathcal{L}_{C \sqsubseteq D^\prime})$
by multiplying a small positive constant $\phi$ that encourages splitting the non-subsumption concepts while does not encourage them to be disjoint. If $\phi=1$, the loss would encourage the non-subsumption concepts to be disjoint. We empirically show that $\phi=1$ produces worse performance as we do not want non-subsumption concepts to be disjoint. 

\begin{table}[t!]
    \resizebox{\textwidth}{!}{
    \begin{minipage}{.47\linewidth}
        \caption{Summary of classes, relations and axioms in different ontologies. NF$_i$ represents the $i^{th}$ normal form.}
        \begin{tabular}{lccc}
        \hline Ontology & GO & GALEN & ANATOMY \\
        \hline
        Classes & 45895 & 24353 & 106363 \\
        Relations & 9 & 1010 & 157 \\
        NF1 & 85480 &  28890 & 122142 \\
        NF2 & 12131 & 13595 & 2121 \\
        NF3 & 20324 & 28118 & 152289 \\
        NF4 & 12129 & 13597 & 2143 \\
        Disjoint & 30 & 0 & 184 \\
        \hline
        \end{tabular}
    \label{tab:dataset}
    \end{minipage}%
    \quad
    \begin{minipage}{.47\linewidth}
      \caption{The accuracies (for which the prediction is true if and only if the subclass box is exactly inside the superclass box) achieved by the embeddings of different approaches in terms of geometric interpretation of the classes in various ontologies.}
        \label{tab:el_syntax_seman}
        \resizebox{\textwidth}{!}{
            \begin{tabular}{ccccccc}
            \hline 
            & ELEm & EmEL$^{++}$ & BoxEL & & \\
            \hline 
            GO & 0.250 & 0.415 & \textbf{0.489 } \\
            GALEN & 0.480 & 0.345 & \textbf{0.788}  \\
            ANATOMY & 0.069 & 0.215 & \textbf{0.453}  \\
            \hline
    \end{tabular}}
    \label{tab:subsumption_accuracy}
    \end{minipage} 
    }
\end{table}



\subsubsection{Datasets} We use three biomedical ontologies 
as our benchmark. 1) \textbf{Gene Ontology (GO)} \cite{harris2004gene} integrates the representation of genes and their functions across all species. 2) \textbf{GALEN} \cite{rector1996galen} is a clinical ontology. 3) \textbf{Anatomy} \cite{mungall2012uberon} is a ontology that represents linkages of different phenotypes to genes. Table \ref{tab:dataset} summarizes the statistical information of these datasets. The subclass relations are split into training set (70\%), validation set (20\%) and testing set (10\%), respectively.

\begin{table*}[t!]
    \centering
    \caption{The ranking based measures of embedding models for sumbsumtion reasoning on the testing set. $*$ denotes the results from \cite{mondala2021emel}.}
      \small
    \begin{tabular}{lllllllll}
    \hline 
    Dataset & Metric & TransE$*$ & TransH$*$ & DistMult$*$ & ELEm & EmEL$^{++}$ & \textbf{BoxEL}  \\
     \hline 
    \multirow{4}{*}{GO} 
    & Hits@10 &  0.00 & 0.00 & 0.00 &  0.09 & 0.10 & 0.03  \\
    & Hits@100 &  0.00 & 0.00 & 0.00  & 0.16 & 0.22 & 0.08 \\
    & AUC &  0.53 & 0.44 & 0.50  & 0.70 & 0.76 & 0.81  \\
    & Mean Rank & - & - & - & 13719 & 11050  & 8980 \\
    \hline 
    \multirow{4}{*}{GALEN} 
    & Hits@10 & 0.00 & 0.00 & 0.00 &  0.07 & 0.10 & 0.02  \\
    & Hits@100 & 0.00 & 0.00 & 0.00 &  0.14 & 0.17 & 0.03  \\
    & AUC &  0.54 & 0.48 & 0.51  & 0.64 & 0.65 & 0.85 \\
    & Mean Rank & - & - & - & 8321 & 8407 & 3584  \\
    \hline 
    \multirow{4}{*}{ANATOMY} 
    & Hits@10 & 0.00 & 0.00 & 0.00 & 0.18 & 0.18 & 0.03  \\
    & Hits@100 &  0.01 & 0.00 & 0.00 & 0.38 & 0.40  & 0.04  \\
    & AUC & 0.53 & 0.44 & 0.49 &  0.73 & 0.76  & 0.91  \\
    & Mean Rank & - & - & - & 28564 & 24421  & 10266  \\
    \hline 
    \end{tabular}
    \label{tab:ranking_measure}
\end{table*}

\subsubsection{Evaluation protocol} 
Two strategies can be used to measure the effectiveness of the embeddings. 
1) Ranking based measures rank the probability of $C$ subsumed by all concepts. We evaluate and report four ranking based measures. Hits@10, Hits@100 describe the fraction of true cases that appear in the first $10$ and $100$ test cases of the sorted rank list, respectively. Mean rank computes the arithmetic mean over all individual ranks (i.e. $\mathrm{MR}=\frac{1}{|\mathcal{I}|} \sum_{\mathrm{rank} \in \mathcal{I}} \mathrm{rank}$, where $rank$ is the individual rank), while AUC computes the area under the ROC curve. 
2) Accuracy based measure is a stricter criterion, for which the prediction is true if and only if the subclass box is exactly inside the superclass box (even not allowing the subclass box slightly outside the superclass box). We use this measure as it evaluates the performance of embeddings on retaining the underlying characteristics of ontology in vector space. We only compare ELEm and EmEL$^{++}$ as KGE baselines fail in this setting (KGEs cannot preserve the ontology). 

\subsubsection{Implementation details} The ontology is normalized into standard normal forms, which comprise a set of axioms that can be used as the \emph{positive samples}. Similar to previous works \cite{kulmanov2019embeddings}, we perform normalization using the OWL APIs and the APIs offered by the jCel reasoner [18]. The hyperparameter for negative sampling is set to $\phi=0.05$. 
For ELEm and EmEL$^{++}$, the embedding size is searched from $n=[50,100,200]$ and margin parameter is searched from $\gamma=[-0.1, 0, 0.1]$. Since box embedding has double the number of parameters of ELEm and EmEL$^{++}$, we search the embedding size from $n=[25,50,100]$ for BoxEL. We summarize the best performing hyperparameters in our supplemental material.
All experiments are evaluated with $10$ random seeds and the mean results are reported for numerical comparisons. 

\subsubsection{Baselines} We compare the state-of-the-art $\mathcal{EL}^{++}$ embeddings (ELEm) \cite{kulmanov2019embeddings}, the first geometric embeddings of $\mathcal{EL}^{++}$, as well as the extension EmEL$^{++}$ \cite{mondala2021emel} that additionally considers the role inclusion and role chain embedding, as our major baselines. For comparison with classical methods, we also include the reported results of three classical KGEs in \cite{mondala2021emel}, including TransE \cite{BordesUGWY13}, TransH \cite{wang2014knowledge} and DistMult \cite{yang2014embedding}. 

\subsubsection{Results} Table~\ref{tab:ranking_measure} summarizes the ranking based measures of embedding models. We first observe that both ELEm and EmEL$^{++}$ perform much better than the three standard KGEs (TransE, TransH, and DistMult) on all three datasets, especially on hits@k for which KGEs fail, showcasing the limitation of KGEs and the benefits of geometric embeddings on encoding logic structures. EmEL$^{++}$ performs slightly better than ELEm on all three datasets. 
Overall, our model BoxEL outperforms ELEm and EmEL$^{++}$. 
In particular, we find that for Mean Rank and AUC, our model achieves significant performance gains on all three datasets.
Note that Mean Rank and AUC have theoretical advantages over hits@k because hits@k is sensitive to any model performance changes while Mean Rank and AUC reflect the average performance, demonstrating that BoxEL achieves better average performance. 
Table~\ref{tab:subsumption_accuracy} shows the accuracies of different embeddings in terms of the geometric interpretation of the classes in various ontologies. 
It clearly demonstrates that BoxEL outperforms ELEm and EmEL$^{++}$ by a large margin, showcasing that BoxEL preserves the underlying ontology characteristics in vector space better than ELEm and EmEL$^{++}$ that use ball embeddings.

\begin{table*}[t!]
    \centering
    \caption{Prediction performance on protein-protein interaction (yeast).}
    \resizebox{\textwidth}{!}{
    \begin{tabular}{ccccccccc}
        \hline 
        Method & \makecell[c]{Raw\\ Hits@10} & \makecell[c]{Filtered\\ Hits@10}  & \makecell[c]{Raw\\ Hits@100} & \makecell[c]{Filtered\\ Hits@100}  & \makecell[c]{Raw\\ Mean Rank}  & \makecell[c]{Filtered\\ Mean Rank}  & \makecell[c]{Raw\\ AUC} & \makecell[c]{Filtered\\ AUC} \\
        \hline 
        TransE    & 0.06 & 0.13 & 0.32 & 0.40	& 1125 & 1075 &	0.82 & 0.83 \\
        BoxE	    & 0.08 & 0.14 & 0.36 & 0.43	& 633 & 620 &	0.85 & 0.85 \\
        SimResnik  & \textbf{0.09} & 0.17 & 0.38 & 0.48	& 758  & 707  &	0.86 & 0.87 \\
        SimLin    & 0.08 & 0.15 & 0.33 & 0.41	& 875  & 825  &	0.8  & 0.85 \\
        ELEm        & 0.08 & 0.17 & 0.44 & 0.62	& 451  & 394  &	0.92 & 0.93 \\
        EmEL$^{++}$      & 0.08 & 0.16 & 0.45 & 0.63 & 451  & 397  & 0.90 & 0.91 \\
        Onto2Vec    & 0.08 & 0.15 & 0.35 & 0.48 & 641  & 588  & 0.79 & 0.80 \\
        OPA2Vec	    & 0.06 & 0.13 &	0.39 & 0.58 & 523  & 467  & 0.87 & 0.88 \\
        BoxEL    & \textbf{0.09} & \textbf{0.20} & \textbf{0.52} & \textbf{0.73} & \textbf{423} & \textbf{379} & \textbf{0.93} & \textbf{0.94} \\
        \hline 
    \end{tabular}}
    \label{tab:ppi_result}
\end{table*}

\begin{table*}[t!]
    \centering
    \caption{Prediction performance on protein-protein interaction (human).}
    \resizebox{\textwidth}{!}{
    \begin{tabular}{ccccccccc}
        \hline 
        Method & \makecell[c]{Raw\\ Hits@10} & \makecell[c]{Filtered\\ Hits@10}  & \makecell[c]{Raw\\ Hits@100} & \makecell[c]{Filtered\\ Hits@100}  & \makecell[c]{Raw\\ Mean Rank}  & \makecell[c]{Filtered\\ Mean Rank}  & \makecell[c]{Raw\\ AUC} & \makecell[c]{Filtered\\ AUC} \\
        \hline 
        TransE	    & 0.05 & \textbf{0.11} & 0.24 & 0.29	& 3960 & 3891 &	0.78 & 0.79 \\
        BoxE	    & 0.05 & 0.10 & 0.26 & 0.32	& 2121 & 2091 &	0.87 & 0.87 \\
        SimResnik	& 0.05 & 0.09 & 0.25 & 0.30	& 1934 & 1864 &	0.88 & 0.89 \\
        SimLin	    & 0.04 & 0.08 & 0.20 & 0.23	& 2288 & 2219 &	0.86 & 0.87 \\
        ELEm        & 0.01 & 0.02 & 0.22 & 0.26	& 1680 & 1638 &	0.90 & 0.90 \\
        EmEL$^{++}$       & 0.01 & 0.03 & 0.23 & 0.26 & 1671 & 1638 & 0.90 & 0.91 \\
        Onto2Vec    & 0.05 & 0.08 & 0.24 & 0.31	& 2435 & 2391 &	0.77 & 0.77 \\
        OPA2Vec	    & 0.03 & 0.07 & 0.23 & 0.26	& 1810 & 1768 &	0.86 & 0.88 \\
        BoxEL (Ours) & \textbf{0.07} & 0.10 & \textbf{0.42} & \textbf{0.63} & \textbf{1574} & \textbf{1530} & \textbf{0.93} & \textbf{0.93} \\
        \hline 
    \end{tabular}}
    \label{tab:ppi_result_human}
\end{table*}

\subsection{Protein-Protein Interactions}

\subsubsection{Dataset} We use a biomedical knowledge graph built by \cite{kulmanov2019embeddings} from Gene Ontology (TBox) and STRING database (ABox) to conduct this task. Gene Ontology contains information about the functions of proteins, while STRING database consists of the protein-protein interactions. We use the protein-protein interaction data of yeast and human organisms, respectively. For each pair of proteins $(P_1, P_2)$ that exists in STRING, we add a role assertion $\rel{interacts}(P1, P2)$. If protein $P$ is associated with the function $F$, we add a membership axiom $\{P\} \sqsubseteq \exists \rel{hasFunction}.F$, the membership assertion can be regarded as a special case of NF3, in which $P$ is a point (i.e. zero-volume box). The interaction pairs of proteins are split into training (80\%), testing (10\%) and validation (10\%) sets. To perform prediction for each protein pair $(P_1, P_2)$, we predict whether the role assertion $\rel{interacts}(P_1, P_2)$ hold. This can be measured by Eq.(\ref{eq:ppi}).
\begin{equation}\label{eq:ppi}
    P(\rel{interacts}(P_1, P_2)) = \left\|T^{\rel{interacts}}_w(m_w(P_1)) - m_w(P_2) \right\|_2.
\end{equation}
where $T^{\rel{interacts}}_w$ is the affine transformation function for relation $\rel{interacts}$. For each positive interaction pair $\rel{interacts}(P_1, P_2)$, we generate a corrupted negative sample by randomly replacing one of the head and tail proteins. 

\subsubsection{Baselines} We consider ELEm \cite{kulmanov2019embeddings} and EmEL$^{++}$ \cite{mondala2021emel} as our two major baselines as they have been shown outperforming the traditional KGEs. We also report the result of Onto2Vec \cite{DBLP:journals/bioinformatics/SmailiGH18} that treats logical axioms as a text corpus and OPA2Vec \cite{smaili2019opa2vec} that combines logical axioms with annotation properties. Besides, we report the results of two semantic similarity measures: Resnik’s similarity and Lin’s similarity in \cite{kulmanov2019embeddings}. For KGEs, we compare TransE \cite{BordesUGWY13}) and BoxE \cite{BoxE}.
We report the hits@10, hits@100, mean rank and AUC (area under the ROC curve) as explained before for numerical comparison. Both raw ranking measures and filtered ranking measures that ignore the triples that are already known to be true in the training stage are reported. Baseline results are taken from the standard benchmark developed by \cite{DBLP:journals/bib/KulmanovSGH21}.\footnote{https://github.com/bio-ontology-research-group/machine-learning-with-ontologies}

\subsubsection{Overall Results} Table~\ref{tab:ppi_result} and Table~\ref{tab:ppi_result_human} summarize the performance of protein-protein prediction in yeast and human organisms, respectively. 
We first observe that similarity based methods (SimResnik and SimLin) roughly outperform TransE, showcasing the limitation of classical knowledge graph embeddings. BoxE roughly outperforms TransE as it does encode some logical properties.  
The geometric methods ELEm and EmEL$^{++}$ fail on the hits@10 measures and does not show significant performance gains on the hits@100 measures in human dataset.
However, ELEm and EmEL$^{++}$ outperform TransE, BoxE and similarity based methods on Mean Rank and AUC by a large margin, especially for the Mean Rank, showcasing the expressiveness of geometric embeddings. Onto2Vec and OPA2Vec achieve relatively better results than TransE and similarity based methods, but cannot compete ELEm and EmEL$^{++}$. We conjecture that this is due to the fact that they mostly consider annotation information but cannot encode the logical structure explicitly. 
Our method, BoxEL consistently outperforms all methods in hits@100, Mean Rank and AUC in both datasets, except the competitive results of hits@10, showcasing the better expressiveness of BoxEL. 

\begin{table}[t!]
    \resizebox{\textwidth}{!}{
    \begin{minipage}{.47\linewidth}
      \caption{The performance of BoxEL with affine transformation (AffineBoxEL) and BoxEL with translation (TransBoxEL) on yeast protein-protein interaction.}
      \centering
      \resizebox{\textwidth}{!}{
      \begin{tabular}{ccccccc}
        \hline 
        Method & \multicolumn{2}{c}{EmEL} & \multicolumn{2}{c}{TransBoxEL} & \multicolumn{2}{c}{AffineBoxEL} \\
        & Raw & Filtered  & Raw & Filtered  & Raw & Filtered \\
        \hline 
        Hits@10 & 0.08 & 0.17  & 0.04 & 0.18 & \textbf{0.09} & \textbf{0.20}\\
        Hits@100 & 0.44 & 0.62 & 0.54 & 0.68 & \textbf{0.52} & \textbf{0.73}\\
        Mean Rank & 451 & 394 & 445 & 390 & \textbf{423} & \textbf{379} \\
        AUC & 0.92 & 0.93 & \textbf{0.93} & 0.93 & \textbf{0.93} & \textbf{0.94} \\
        \hline 
    \end{tabular}}
    \label{tab:transform_translation}
    \end{minipage}%
    \quad
    \begin{minipage}{.47\linewidth}
      \centering
        \caption{The performance of BoxEL with point entity embedding and box entity embedding on yeast protein-protein interaction dataset.}
        \resizebox{\textwidth}{!}{
        \begin{tabular}{ccccccc}
        \hline 
        Method & \multicolumn{2}{c}{EmEL} & \multicolumn{2}{c}{BoxEL (boxes)} & \multicolumn{2}{c}{BoxEL (points)} \\
        & Raw & Filtered  & Raw & Filtered  & Raw & Filtered \\
        \hline 
        Hits@10 & 0.08 & 0.17  & \textbf{0.09} & 0.19 & \textbf{0.09} & \textbf{0.20}\\
        Hits@100 & 0.44 & 0.62 & 0.48 & 0.68 &  \textbf{0.52} & \textbf{0.73}\\
        Mean Rank & 451 & 394 & 450 & 388 & \textbf{423} & \textbf{379} \\
        AUC & 0.92 & 0.93 & 0.92 & 0.93 & \textbf{0.93} & \textbf{0.94} \\
        \hline 
    \end{tabular}}
    \label{tab:points_boxes}
    \end{minipage} 
    }
\end{table}

\subsection{Ablation Studies} 

\subsubsection{Transformation vs Translation} To study the contributions of using boxes for modeling concepts and using affine transformation for modeling relations, we conduct an ablation study by comparing relation embeddings with affine transformation (AffineBoxEL) and translation (TransBoxEL). The only difference of TransBox to the AffineBox is that TransBox does not associate a scaling factor for each relation. Table~\ref{tab:transform_translation} clearly shows that TransBoxEL outperforms EmEL$^{++}$, showcasing the benefits of box modeling compared with ball modeling. While AffineBoxEL further improves TransBoxEL, demonstrating the advantages of affine transformation. 
Hence, we could conclude that both of our proposed entity and relation embedding components boost the performance.

\subsubsection{Entities as Points vs Boxes} As mentioned before, distinguishing entities and concepts by identifying entities as points has better theoretical properties. Here, we study how this distinction influences the performance. For this purpose, we eliminate the ABox axioms by replacing each individual with a singleton class and rewriting relation assertions $r(a, b)$ and concept assertions $C(a)$ as $\{a\} \sqsubseteq \exists r.\{b\}$ and $\{a\} \sqsubseteq C$, respectively. In this case, we only have TBox embeddings and the entities are embedded as regular boxes. Table~\ref{tab:points_boxes} shows that for hits@k, there is marginal significant improvement of point entity embedding over boxes entity embedding, however, point entity embedding consistently outperforms box entity embedding on Mean Rank and AUC, showcasing the benefits of distinguishing entities and concepts.

\section{Conclusion}

This paper proposes BoxEL, a geometric KB embedding method that explicitly models the logical structure expressed by the theories of $\mathcal{EL}^{++}$. 
Different from the standard KGEs that simply ignore the analytical guarantees, BoxEL provides \textit{soundness} guarantee for the underlying logical structure by incorporating background knowledge into machine learning tasks, offering a more reliable and logic-preserved fashion for KB reasoning. 
The empirical results further demonstrate that BoxEL outperforms previous KGEs and $\mathcal{EL}^{++}$ embedding approaches on subsumption reasoning over three ontologies and predicting protein-protein interactions in a real-world biomedical KB. For future work, since the volume of boxes can naturally capture joint probability distribution \cite{vilnis2018probabilistic}, we are planning to extend the work to probabilistic description logic StatisticalEL \cite{penaloza2017towards} that allows making statements about statistical proportions in a population.


\paragraph*{Supplemental Material Statement:} Source code and datasets are available for reproducing the results.\footnote{https://github.com/Box-EL/BoxEL}
Full proofs of propositions and lemmas are in the Appendix.


\section*{Acknowledgments}
The authors thank the International Max Planck Research School for Intelligent Systems (IMPRS-IS) for supporting Bo Xiong. This project has received funding from the European Union’s Horizon 2020 research and innovation programme under the Marie Skłodowska-Curie grant agreement No: 860801. 
Nico Potyka was partially funded by DFG projects Evowipe/COFFEE.

\bibliographystyle{splncs04}
\bibliography{reference}

\begin{thebibliography}{10}
\providecommand{\url}[1]{\texttt{#1}}
\providecommand{\urlprefix}{URL }
\providecommand{\doi}[1]{https://doi.org/#1}

\bibitem{BoxE}
Abboud, R., Ceylan, {\.I}.{\.I}., Lukasiewicz, T., Salvatori, T.: Boxe: {A} box
  embedding model for knowledge base completion. In: NeurIPS (2020)

\bibitem{baader2005pushing}
Baader, F., Brandt, S., Lutz, C.: Pushing the el envelope. In: IJCAI. vol.~5,
  pp. 364--369 (2005)

\bibitem{baader2003description}
Baader, F., Calvanese, D., McGuinness, D., Patel-Schneider, P., Nardi, D.,
  et~al.: The description logic handbook: Theory, implementation and
  applications. Cambridge university press (2003)

\bibitem{BordesUGWY13}
Bordes, A., Usunier, N., Garc{\'{\i}}a{-}Dur{\'{a}}n, A., Weston, J.,
  Yakhnenko, O.: Translating embeddings for modeling multi-relational data. In:
  {NIPS}. pp. 2787--2795 (2013)

\bibitem{gene2015gene}
Consortium, G.O.: Gene ontology consortium: going forward. Nucleic acids
  research  \textbf{43}(D1),  D1049--D1056 (2015)

\bibitem{dasgupta2020improving}
Dasgupta, S.S., Boratko, M., Zhang, D., Vilnis, L., Li, X., McCallum, A.:
  Improving local identifiability in probabilistic box embeddings. In: NeurIPS
  (2020)

\bibitem{DettmersMS018}
Dettmers, T., Minervini, P., Stenetorp, P., Riedel, S.: Convolutional 2d
  knowledge graph embeddings. In: {AAAI}. pp. 1811--1818. {AAAI} Press (2018)

\bibitem{DBLP:books/daglib/0006106}
G{\"{a}}rdenfors, P.: Conceptual spaces - the geometry of thought. {MIT} Press
  (2000)

\bibitem{garg2019quantum}
Garg, D., Ikbal, S., Srivastava, S.K., Vishwakarma, H., Karanam, H.P.,
  Subramaniam, L.V.: Quantum embedding of knowledge for reasoning. In: NeurIPS.
  pp. 5595--5605 (2019)

\bibitem{graua2008web}
Graua, B.C., Horrocksa, I., Motika, B., Parsiab, B., Patel-Schneiderc, P.,
  Sattlerb, U.: Web semantics: Science, services and agents on the world wide
  web. Web Semantics: Science, Services and Agents on the World Wide Web
  \textbf{6},  309--322 (2008)

\bibitem{DBLP:conf/kr/Gutierrez-Basulto18}
Guti{\'{e}}rrez{-}Basulto, V., Schockaert, S.: From knowledge graph embedding
  to ontology embedding? an analysis of the compatibility between vector space
  representations and rules. In: {KR}. pp. 379--388. {AAAI} Press (2018)

\bibitem{harris2004gene}
Harris, M., Clark, J., Ireland, A., Lomax, J., Ashburner, M., Foulger, R.,
  Eilbeck, K., Lewis, S., Marshall, B., Mungall, C., et~al.: The gene ontology
  (go) database and informatics resource nucleic acids research, 32. D258--D261
   (2004)

\bibitem{DBLP:journals/jar/KazakovKS14}
Kazakov, Y., Kr{\"{o}}tzsch, M., Simancik, F.: The incredible {ELK} - from
  polynomial procedures to efficient reasoning with el ontologies. J. Autom.
  Reason.  \textbf{53}(1),  1--61 (2014)

\bibitem{DBLP:journals/corr/KingmaB14}
Kingma, D.P., Ba, J.: Adam: {A} method for stochastic optimization. In: {ICLR}
  (Poster) (2015)

\bibitem{kulmanov2019embeddings}
Kulmanov, M., Liu{-}Wei, W., Yan, Y., Hoehndorf, R.: {EL} embeddings: Geometric
  construction of models for the description logic {EL++}. In: {IJCAI}. pp.
  6103--6109. ijcai.org (2019)

\bibitem{DBLP:journals/bib/KulmanovSGH21}
Kulmanov, M., Smaili, F.Z., Gao, X., Hoehndorf, R.: Semantic similarity and
  machine learning with ontologies. Briefings Bioinform.  \textbf{22}(4) (2021)

\bibitem{li2018smoothing}
Li, X., Vilnis, L., Zhang, D., Boratko, M., McCallum, A.: Smoothing the
  geometry of probabilistic box embeddings. In: {ICLR}. OpenReview.net (2019)

\bibitem{DBLP:conf/aaai/LinLSLZ15}
Lin, Y., Liu, Z., Sun, M., Liu, Y., Zhu, X.: Learning entity and relation
  embeddings for knowledge graph completion. In: {AAAI}. pp. 2181--2187. {AAAI}
  Press (2015)

\bibitem{DBLP:conf/icml/LiuWY17}
Liu, H., Wu, Y., Yang, Y.: Analogical inference for multi-relational
  embeddings. In: {ICML}. Proceedings of Machine Learning Research, vol.~70,
  pp. 2168--2178. {PMLR} (2017)

\bibitem{mondala2021emel}
Mondal, S., Bhatia, S., Mutharaju, R.: Emel++: Embeddings for {EL++}
  description logic. In: {AAAI} Spring Symposium: Combining Machine Learning
  with Knowledge Engineering. {CEUR} Workshop Proceedings, vol.~2846.
  CEUR-WS.org (2021)

\bibitem{mungall2012uberon}
Mungall, C.J., Torniai, C., Gkoutos, G.V., Lewis, S.E., Haendel, M.A.: Uberon,
  an integrative multi-species anatomy ontology. Genome biology
  \textbf{13}(1),  1--20 (2012)

\bibitem{DBLP:conf/icml/NickelTK11}
Nickel, M., Tresp, V., Kriegel, H.: A three-way model for collective learning
  on multi-relational data. In: {ICML}. pp. 809--816. Omnipress (2011)

\bibitem{ozccep2020cone}
{\"{O}}z{\c{c}}ep, {\"{O}}.L., Leemhuis, M., Wolter, D.: Cone semantics for
  logics with negation. In: {IJCAI}. pp. 1820--1826. ijcai.org (2020)

\bibitem{patel2020representing}
Patel, D., Dasgupta, S.S., Boratko, M., Li, X., Vilnis, L., McCallum, A.:
  Representing joint hierarchies with box embeddings. In: Automated Knowledge
  Base Construction (2020), \url{https://openreview.net/forum?id=J246NSqR_l}

\bibitem{penaloza2017towards}
Pe{\~n}aloza, R., Potyka, N.: Towards statistical reasoning in description
  logics over finite domains. In: International Conference on Scalable
  Uncertainty Management (SUM). pp. 280--294. Springer (2017)

\bibitem{rector1996galen}
Rector, A.L., Rogers, J.E., Pole, P.: The galen high level ontology. In:
  Medical Informatics Europe’96, pp. 174--178. IOS Press (1996)

\bibitem{ren2020query2box}
Ren, H., Hu, W., Leskovec, J.: Query2box: Reasoning over knowledge graphs in
  vector space using box embeddings. In: {ICLR}. OpenReview.net (2020)

\bibitem{betae}
Ren, H., Leskovec, J.: Beta embeddings for multi-hop logical reasoning in
  knowledge graphs. In: Neurips (2020)

\bibitem{DBLP:journals/bioinformatics/SmailiGH18}
Smaili, F.Z., Gao, X., Hoehndorf, R.: Onto2vec: joint vector-based
  representation of biological entities and their ontology-based annotations.
  Bioinform.  \textbf{34}(13),  i52--i60 (2018)

\bibitem{smaili2019opa2vec}
Smaili, F.Z., Gao, X., Hoehndorf, R.: Opa2vec: combining formal and informal
  content of biomedical ontologies to improve similarity-based prediction.
  Bioinform.  \textbf{35}(12),  2133--2140 (2019)

\bibitem{DBLP:journals/ws/SteigmillerLG14}
Steigmiller, A., Liebig, T., Glimm, B.: Konclude: System description. J. Web
  Semant.  \textbf{27-28},  78--85 (2014)

\bibitem{TrouillonWRGB16}
Trouillon, T., Welbl, J., Riedel, S., Gaussier, {\'{E}}., Bouchard, G.: Complex
  embeddings for simple link prediction. In: {ICML}. {JMLR} Workshop and
  Conference Proceedings, vol.~48, pp. 2071--2080. JMLR.org (2016)

\bibitem{vilnis2018probabilistic}
Vilnis, L., Li, X., Murty, S., McCallum, A.: Probabilistic embedding of
  knowledge graphs with box lattice measures. In: {ACL} {(1)}. pp. 263--272.
  Association for Computational Linguistics (2018)

\bibitem{wang2017knowledge}
Wang, Q., Mao, Z., Wang, B., Guo, L.: Knowledge graph embedding: A survey of
  approaches and applications. IEEE Transactions on Knowledge and Data
  Engineering  \textbf{29}(12),  2724--2743 (2017)

\bibitem{wang2014knowledge}
Wang, Z., Zhang, J., Feng, J., Chen, Z.: Knowledge graph embedding by
  translating on hyperplanes. In: {AAAI}. pp. 1112--1119. {AAAI} Press (2014)

\bibitem{yang2014embedding}
Yang, B., Yih, W., He, X., Gao, J., Deng, L.: Embedding entities and relations
  for learning and inference in knowledge bases. In: {ICLR} (Poster) (2015)

\end{thebibliography}

\section*{Proofs of Theorems}

\begin{proposition}\label{prop:1}
We have 
\begin{enumerate}
    \item If $\mathcal{L}_{C(a)}(w) = 0$, then $\interpw \models C(a)$,
    \item If $\mathcal{L}_{r(a,b)}(w)=0$, then $\interpw \models r(a,b)$.
\end{enumerate}
\end{proposition}
Proposition \ref{prop:1} follows directly from the definitions.

\begin{lemma}
\label{contains_lemma}
\begin{enumerate}
    \item $0 \leq \contains(B_1, B_2) \leq 1$,
    \item $\contains(B_1, B_2) = 0$ implies $B_1 \subseteq B_2$,
    \item $\contains(B_1, B_2) = 1$ implies $B_1 \cap B_2 = \emptyset$.
\end{enumerate}
\end{lemma}
\begin{proof}
1. Since $B_1 \cap B_2 \subseteq B_1$, $\frac{\Vol(B_1 \cap B_2)}{\Vol(B_1)}\leq 1$.
The modified volume is also non-negative, so that the fraction is non-negative and
$0 \leq \contains(B_1, B_2) \leq 1$.

2. If $\contains(B_1, B_2) = 0$ then we must have $\Vol(B_1 \cap B_2)=1$ and
therefore $\Vol(B_1 \cap B_2) = \Vol(B_1) \neq 0$.
Since $B_1 \cap B_2 \subseteq B_1$, this is only possible if $B_1 \cap B_2 = B_1$,
but this implies that $B_1 \subseteq B_2$.

3. If $\contains(B_1, B_2) = 1$, we must have $\Vol(B_1 \cap B_2)=0$. By definition
of the modified volume this is only possible if $B_1 \cap B_2 = \emptyset$.
\end{proof}

\begin{proposition}
If $\mathcal{L}_{C \sqsubseteq D}(w)=0$, 
then $\interpw \models C \sqsubseteq D$,
where we exclude the inconsistent case $C=\{a\}, D=\bot$.
\end{proposition}
\begin{proof}
For $D\neq \bot$, the claim follows from Lemma \ref{contains_lemma}. 
For $D=\bot$, $\mathcal{L}_{C \sqsubseteq \bot}(w)=0$ implies that 
$\Box_w(C) = \emptyset$ and the claim is trivially true.
\end{proof}

\begin{proposition}
If $\mathcal{L}_{C \sqcap D \sqsubseteq E}(w)=0$,
then $\interpw \models C \sqcap D \sqsubseteq E$,
where we exclude the inconsistent case ${a} \sqcap {a} \sqsubseteq \bot$ (that is, $C=D=\{a\}, E=\bot$).
\end{proposition}
\begin{proof}
We have to show that for every $x \in \Box_w(C)$, there is 
a $y \in \Box_w(D)$ such that $T^r_w(x) = y$. Note that
$\contains(T^r_w(\Box_w(C)), \Box_w(D))=0$
implies that $T^r_w(\Box_w(C)) \subseteq \Box_w(D)$
according to Lemma \ref{contains_lemma}. 
Since $x \in \Box_w(C)$,
we have $T^r_w(x) = y \in \Box_w(D)$.
\end{proof}

\begin{proposition}
If $\mathcal{L}_{C \sqsubseteq \exists r.D}(w) = 0$, 
then $\interpw \models C \sqsubseteq \exists r.D$.
\end{proposition}
The proof of Proposition 4 is analogous to the proof of Proposition 3.

\begin{proposition}
If $\mathcal{L}_{\exists r.C \sqsubseteq D }(w) = 0$, 
then $\interpw \models \exists r.C \sqsubseteq D$.
\end{proposition}
\begin{proof}
We have to show that if $T^{-r}_w(x)=y$ and
$y \in \Box_w(C)$, then $x \in \Box_w(D)$. $\contains(T^{-r}_w(\Box_w(C)), \Box_w(D))=0$
implies that $T^{-r}_w(\Box_w(C)) \subseteq \Box_w(D)$
according to Lemma \ref{contains_lemma}.
Since $y \in \Box_w(C)$, we have 
$x = T^{-r}_w(y) \in \Box_w(D)$.
\end{proof}

\section*{Hyperparameters}
We only search the hyperparameters for ElEm, EmEL$^{++}$ and BoxEL. Table \ref{tab:hyperparameters} summarizes the hyperparameters for each model on different datasets. 

\begin{table}[]
    \centering
    \caption{Best performing hyperparameters for each model, where $\gamma$ is the margin loss parameter and $n$ is the dimension.
}
    \begin{tabular}{rrrrrrr}
        \hline & \multicolumn{2}{c}{EmEL$^{++}$} & & \multicolumn{2}{c}{ ElEm } & BoxEL  \\
        \cline {2-3} \cline {5-6} & $n$ & $\gamma$ & & $n$ & $\gamma$ & $\gamma$ \\
        \hline 
        GO & 100 & $-0.1$ & & 100 & $-0.1$  & 50\\
        GALEN & 50 & $0.0$ & & 50 & $0.0$ & 25 \\
        ANATOMY & 200 & $-0.1$ & & 200 & $-0.1$ & 100  \\
        Yeast & 100 & $-0.1$ & & 100 & $-0.1$ & 50 \\
        Human & 200 & $-0.1$ & & 200 & $-0.1$ & 50 \\
        \hline 
    \end{tabular}
    \label{tab:hyperparameters}
\end{table}

\end{document}